\documentclass{article}

\title{Revisiting the DARPA Communicator Data using Conversation Analysis}

\author{Peter Wallis \\
The University of Sheffield, Department of Computer Science,\\
Regent Court, 211 Portobello Street, Sheffield, S1 4DP, UK\\
pwallis@acm.org
}

\date{}
\begin{document}
\maketitle
{\bf Abstract:}
The state of the art in human computer conversation leaves something
to be desired and, indeed, talking to a computer can be
down-right annoying.  This paper describes an approach to identifying
``opportunities for improvement'' in these systems by looking for abuse
in the form of swear words.   The premise is that humans swear at
computers as a sanction and, as such, swear words represent a
point of failure where the system did not behave as it should.
Having identified \emph{where} things went wrong, we can work
backward through the transcripts and, using conversation analysis
(CA) work out \emph{how} things went wrong.
Conversation analysis is a qualitative methodology and can
appear quite alien - indeed unscientific - to those of us from a
quantitative background.  The paper starts with a description of
Conversation analysis in its modern form, and then goes on to
apply the methodology to transcripts of frustrated and annoyed
users in the DARPA Communicator project.
The conclusion is that there is at least one species of failure
caused by the inability of the Communicator systems to handle mixed
initiative at the discourse structure level.
Along the way, I hope to demonstrate that there is an alternative
future for computational linguistics that does not rely on
larger and larger text corpora.

\vspace{10mm}


\vspace{10mm}

{\bf Bio:}
Peter Wallis is currently at Sheffield University as part of the 13M Euro
Companions integration project, and a member of the European HUMAINE NoE.
He gained his doctorate in 1995 from RMIT University working with the
Melbourne based SIM Group on semantic information retrieval, and was technical
lead on a very successful information extraction system based on a library of
`fact extractors'.  Having done post graduate studies in science leadership,
experience with defence research, two SMEs, and lecturing software project
management (among other things) he has considerable experience working on the
boundary where academia interfaces with innovation.  Since 2000 his research
interests have focused on social skills for conversational agents, and more
recently, the affective nature of human-robot interaction.
\pagebreak

\section{Introduction}
Why do people swear at computers?
Verbal abuse of conversational machines is more wide-spread and more
consistent than one might think.  De Angeli and Carpenter~\cite{deAngeli05}
looked at logs for the Jabberwacky chat-bot~\cite{jabberwacky} and found that
10\% of the \emph{words} used by humans talking to the system were related to
insulting or offensive language.  Indeed ``the f word'' ranked ninth in a
list of the open class word stems.  Naturally Jabberwacky is a system
that is there to entertain, and one might expect that people are more
sensible when they use a spoken language system to perform some useful
task.  They certainly are better behaved under experimental conditions
but it is not clear they are well behaved when they think no-one is
looking.  Circumstantial evidence suggests that some systems `bring on'
verbal abuse.
Several telephone based services in Australia use automatic speech recognition
to implement automated call handling including ordering a taxi, betting on
the horses, telephone directory assistance, and paying for use of
Melbourne's CityLink toll-way.  In 2002 circumstantial evidence suggested
people were happy using the taxi service but often ended up swearing
at the CityLink system.  What is it about some systems that upsets
users and, indeed, are they really upset or just `playing' in some way?
Commercial interests mean it is difficult for academics to get access
to real data without some form of non-disclosure agreement, however
unpublished results~\cite{Webb07} from the AMITI\`{E}S project~\cite{amities04} suggest
that just 2\% of calls resulted in a recognisably annoyed caller.  On the
surface this appears quite good.  If however one in fifty callers are changing
bank so that they can talk to a human, then perhaps there is an actual
commercial reason to look at why people swear at machines.
In the work described here, I look at data from the DARPA
Communicator project~\cite{CommEval} in which high profile research
institutions looked at the nature of human-computer conversation in
the context of making travel plans. 
With these state-of-the-art systems, over a quarter of the calls resulted in
the user \emph{not} wanting to ``use the system on a regular basis''.
The purpose of this paper is to show how conversation
analysis~\cite{sacks,HutWoo98}, or CA, can help.

Current best practice in the field would use statistical
analysis of the Communicator transcripts to make generally applicable
claims about human-machine conversations.
However, such statistical approaches can only say things about `average'
behaviour.  If, indeed, just 2\% of callers are having problems, then
statistical techniques are unlikely to `notice' any issues.
Conversation analysis has its roots in sociology of the late
1960's and early 70's and the work of Sacks, Schegloff, and
Jefferson~\cite{SSJ74,sacks} and was introduced as a way to study
what people actually \emph{do} in conversation.
Conversation analysis is a qualitative rather than quantitative approach
and the use of CA is quite a paradigm shift for
those of us familiar with computer based language processing.  The
methodology is discussed further below.

So why do people swear at machines? In the de Agneli and Carpenter
paper cited above, it was argued that verbal abuse occurs as a means
of expressing power relations.
The premise being that humans do indeed treat computers as social
actors~\cite{ReeNas96} and, as a social actor, the machine has to ``know its
place'' and act accordingly.  Importantly, ``its place'' is not the same as
that of the person who creates it~\cite{ang01}.   A complimentary view is that
abuse is part of socialization - part of the mechanism that creates
and maintains normative behaviour in human societies~\cite{wall05-abuse}.
Social actors are abused only when they act outside ``their place'' and the
purpose of the abuse in an evolutionary sense (in the same way as the
purpose of eyes is seeing) is to maintain social order. 
It turns out that such a mechanism is described in the CA literature.
Seedhouse~\cite{Seedhouse04} describes how the response of a conversational
partner is either \textbf{seen but un-noticed, is accounted for}, or
\textbf{risks sanction}.
Normal talk in interaction goes \textbf{seen but un-noticed} with greetings
followed by greetings; questions followed by answers and so on.
The response in such an adjacency pair is the normal response and, at most,
transfers information if the response is remembered at all.
When someone says something that is out of the ordinary or provides a
dis-preferred response (see below) then their actions are \textbf{accounted
for}.  If a question is followed by a question, then the asker of the first
questions assumes that the answer to their question must in some way depend
on how they answer the follow-up question.
Finally, If a speaker's actions cannot not be accounted for, then the
speaker's conversational partner will impose \textbf{sanctions}.
Those sanctions, when the conversational partner is a computer and no-one
is looking, take the form of verbal abuse.

The premise is that people verbally abuse dialogue systems,
not when things go wrong, but when the system fails to account for its
non-normal behaviour.  In the work described here, the methodology is
to go through the Communicator transcripts looking for the sanctions - in
the first instance, looking for the swear words - and then work backward to
see how the computer system might have accounted for its actions.

\subsection{A Note on Methodology}

In the work described here, only two conversations are discussed in detail
and this is likely to appear completely unscientific to those of us
familiar with the quantitative approaches involving samples, populations,
and measures of significance.
The methods of corpus based linguistics that dominate computer based
language processing are classically quantitative and typically involve
megabytes of text representing some genre of naturally occurring text
or speech.  This sample and population approach is, in Alasuutari's terms
(cf. ten Have, 1998, page 50) a `factist' approach with the
sample being seen as representative of a population which is, in itself,
inaccessible to the researcher.  The qualitative `specimen' approach
by contrast sees the sample as an \emph{instance} of a class of
thing, to which the researcher has direct access.   To use ten
Have's example, a researcher might set out to learn something about
the life of sparrows by observing a few specimen sparrows. 
Because the specimens \emph{are} sparrows, the what-is-learnt is
\emph{of} sparrows.  One sets out to define a category of thing called
SPARROW by looking at a sparrow. and having a tentative definition of SPARROW,
one can explore the boundaries of the category by examining other sparrows.
Rather than defining `the average sparrow' in terms of a representative
sample, we look for a definition of SPARROW that matches the things we humans,
using our common sense, identify as sparrows.  Note how this `specimen'
approach affects the idea of sampling.
Rather than a random sample of statistically significant size, one would
like `maximum variation' in the sample.  Given a category of thing, SPARROW,
we can define the boundaries of the category by looking at examples on the
edge.  An apparent sparrow that grows
very large and lays its eggs in other birds nests we might decide is not
a SPARROW.  The qualities of the bird that make it not a sparrow would
be worked into our definition of SPARROW.  
We are looking for qualitative differences that distinguish categories rather
than quantitative differences between abstract `populations'. 
One might feel that the researcher is introducing a bias by making the initial
claim that ``this is a sparrow'' however the same bias is needed to say ``here
is a set of sparrows'' to be used as the sample in the quantitative
sample and population approach.  Neither is more objective in principle; both
force theory to be assessed against reality but neither start from an
objective base.
In this paper a small number of Communicator transcripts are selected
and studied in depth.  The science is to identify a linguistic
phenomena that is a `species' of failing in the the Communicator
systems, and then explore the phenomena in more detail. 
Finally, it should be noted that CA is not the only ethnomethodological
approach to language analysis and Eggins and Slade~\cite{EggSla97} group
Dell Hymes (credited with asking `who says what to whom, when, where, why
and how?') and the sociolinguists under the EM banner.
The next section takes a closer look at CA as a methodology with
Section~\ref{communicator} introducing the Communicator data and
applying CA to some of the dialogues found there. Section~\ref{RMI}
discusses the notion of mixed initiative dialogue and shows how the
Communicator systems studied did not provide it.

\section{Conversation Analysis as Ethnomethodology}

In 2000/2001, while working on a large scale embodied conversational agent
(ECA) project~\cite{wmod01}, we surveyed and abandoned a number of approaches
to language analysis including conversation analysis.
At that time CA appeared to be too `low level' to be useful.
Classic CA is very focused on \textbf{turn taking} and Hutchby and
Wooffitt in their chapter on the foundations of conversation analysis
spend eight pages on adjacency pairs and preference (see below) but thirteen
pages discussing the findings of CA on turn taking, overlap and
`interruptions', and another ten pages on turn taking and repair.
Working only with that which is in the text, and banning all talk of
mental attitudes, CA looked very `behaviourist' and this did not fit well
with our cognitivist outlook.
Seedhouse~\cite[p46]{Seedhouse04} sees this as the linguists
misunderstanding, and explains that the right way to interpret the admittedly
vague early writings on CA~\cite{HutWoo98} is via ethnomethodology.

Cultural anthropology has been studying the common sense of others for
centuries and the term `ethnomethodology' (EM) was coined by Harold
Garfinkel~\cite{garf67} to mean a set of techniques that researchers use to
get an `inside view' from members of a community.
Rather than looking at behaviour through the lens
of a researcher's theory, the aim is to understand the workings of the
community from the perspective of community members, using the common
sense reasoning of the members themselves.
 The core of ethnomethodological studies is
to empirically study ``the sociological reasoning of members of a community,
providing explanations of events in terms of the members own
methods''~\cite[p20]{tHave99}.  These methods are just common sense
\emph{within} the community of practice, but can appear quite
bizarre to outsiders.
In a 1984 Herzog film~\cite{Herzog84} Aboriginal elders explain to a mining
engineer
that they don't want mining there because that is ``where the green ants
dream''.   This makes perfect sense from their perspective, but no
sense what so ever to the mining engineer.  
The ultimate aim of ethnomethodologically inspired research is to understand
how members' communities work, not in terms of some external theory, but
in terms of members own commonsense practices. 
Applied to language, we want to model the decisions people make when
they say what they say.  Rather than a mathematical model of word order,
we want to model language production from the perspective of the participants.
The shape of fern leafs might have a beautiful mathematical description but
to write that description is mathematics, not biology.  Similarly, the
issue here is language in use, and the object of study is the community of
practice.

Cultural anthropology classically studied other cultures and noted the
apparently non-sensical behaviour of community members.  Studying
one's own culture is harder because, well, it is just commonsense.
Several techniques have been used including Garfinkel's infamous breaching
experiments in which the researcher deliberately does not behave in accordance
with established norms, often causing the involuntary subjects - the people
around him or her - to become upset and angry.  Another approach that was 
used in our 2001 ECA project~\cite{wmod01} is for the researcher to
take on the role of an outsider who wants to learn what to do.  This gives
the researcher permission to ask na\"{i}ve questions of the community of interest. 

Conversation analysis involves
recording events in their raw form, transcribing what happens and, through
this process, deriving an understanding from the participants' point of
view.  The researcher uses his or her commonsense - not as a researcher,
but as a member of the community of practice - to give a detailed account of
``why this action, in this way, right here''~\cite{Seedhouse04}. 
Hutchby and Wooffitt~\cite{HutWoo98} summarise the insights which form the
methodological basis of CA as follows:
\begin{itemize}
\item Talk-in-interaction is systematically organised and deeply ordered.
\item The production of talk-in-interaction is methodic.
\item The analysis of talk-in-interaction should be based on naturally occurring data.
\item Analysis should not initially be constrained by prior theoretical assumptions.
\end{itemize}
The characterising assumption is that people mostly say things for a reason,
and if not for a reason, then what they say, says something about them to the
people around them.  It is only communication if another language user can
interpret that which is said, and the invisible is irrelevant to
language in use.  The researcher, as an expert language user, has
direct access to the salient features of language as a communication tool.
In general, no detail of talk-in-interaction is too small to ignore, although
not every detail necessarily forms part of an analysis.

\subsection{Using Conversation Analysis}

Paul ten Have in his book ``Doing conversation analysis''~\cite{tHave99}
provides this general outline of the process:
\begin{enumerate}
\item find or make recordings of natural interaction
\item transcribe all or part of the tapes
\item analyse selected episodes
\item report results
\end{enumerate}
The process, ten Have points out, needs to be a re-entrant process
that spirals toward a true understanding.  In early work on CA, the
transcript was seen as a representation of what \emph{actually} happened
uncontaminated by the researcher's theories.  In this re-entrant
process, the researcher can be influenced by theory (which is inevitable)
but the process forces the researcher to pay close attention to the actual
data.  One might start with a wrong theory but, according to the experts, a wrong
theory will not stand up for long in the face of real data.  This intimate
relation between theory and data means that the research should do his or her
own transcription, forcing close attention to the actual recordings.
This process also means that the transcriptions are not intended to capture
everything.
The classic transcription scheme is to use Gail Jefferson's system as presented
in Figure~\ref{GJ}.  Linguists certainly have transcription schemes that
are better at capturing what is said, but they tend to be less readable. In any
case, the point in CA is to capture the \emph{important} things in talk.
The important things are primarily those that influence the participants
commonsense understanding of the interaction, however much of what is said
is `seen but unnoticed.' The researcher's job is to note the everyday,
and show how it contributes to commonsense understanding.

\begin{table}
\begin{tabular}{|rp{30mm}l|}
\hline
(0.5)	&\multicolumn{2}{p{130mm}|}{The number in brackets indicates a time gap in tenths of a second.}\\
(.)	&\multicolumn{2}{p{130mm}|}{A dot enclosed in a bracket indicates a pause in the talk of less than two-tenths of a second.}\\
=	&\multicolumn{2}{p{130mm}|}{The `equals' sign indicates `latching' between utterances. For example:}\\
	&&		S1: yeah September[seventy six=\\
	&&		S2: [September\\
	&&		S1: =it would be\\
	&&		S2: yeah that's right\\
\protect[ ]	&\multicolumn{2}{p{130mm}|}{Square brackets between adjacent lines of concurrent speech indicate the onset and end of a spate of overlapping talk.}\\
.hh	&\multicolumn{2}{p{130mm}|}{A dot before a `h' indicates speaker in-breath. The more h's the longer the in-breath.}\\
hh	&\multicolumn{2}{p{130mm}|}{An `h' indicates an out-breath. The more h's the longer the breath.}\\
(( ))	&\multicolumn{2}{p{130mm}|}{A description enclosed in double bracket indicates a non-verbal activity.  For example ((banging sound)).  Alternatively double brackets may enclose the transcriber's comments on contextual or other features.}\\
-	&\multicolumn{2}{p{130mm}|}{A dash indicates the sharp cut-off of the prior word or sound}\\
:	&\multicolumn{2}{p{130mm}|}{Colons indicate that the speaker has stretched the preceding sound or letter.  The more colons the greater the extent of the stretching.}\\
!	&\multicolumn{2}{p{130mm}|}{Exclamation marks are used to indicate an animated or emphatic tone.}\\
( )	&\multicolumn{2}{p{130mm}|}{Empty parenthesis indicate the presence of an unclear fragment on the tape.}\\
(guess)	&\multicolumn{2}{p{130mm}|}{The words within a single bracket indicate the transcriber's best guess at an unclear utterance.}\\
.	&\multicolumn{2}{p{130mm}|}{A full stop indicates a stopping fall in tone. It does not necessarily indicate the end of a sentence.}\\
,	&\multicolumn{2}{p{130mm}|}{A comma indicates a `continuing' intonation.}\\
?	&\multicolumn{2}{p{130mm}|}{A question mark indicates a rising inflection. It does not necessarily indicate a question.}\\
$\ast$	&\multicolumn{2}{p{130mm}|}{An asterisk indicates a `croaky' pronunciation of the immediately following section.}\\
$\downarrow \uparrow$	&\multicolumn{2}{p{130mm}|}{Pointed arrows indicate a marked falling or rising intonational shift. They are placed immediately before the onset of the shift.}\\
\underline{a}:	&\multicolumn{2}{p{130mm}|}{Less marked falls in pitch can be indicated by using underlining immediately preceding a colon:}\\
	&& S: we (.) really didn't have a lot'v ch\underline{a}:nge \\
a:	&\multicolumn{2}{p{130mm}|}{Less marked rises in pitch can be indicated using a colon which itself is underlined:}\\
	&& J: I hava a red shi\underline{:}rt, \\
\underline{Under}	&\multicolumn{2}{p{130mm}|}{Underlined fragments indicate speaker emphasis.}\\
CAPS	&\multicolumn{2}{p{130mm}|}{Words in capitals mark a section of speech noticeably louder than that surrounding it.}\\
$\circ  ... \circ$	&\multicolumn{2}{p{130mm}|}{Degree signs are used to indicate that the talk they encompass is spoken noticeably quieter than the surrounding talk.}\\
Thaght	&\multicolumn{2}{p{130mm}|}{A `gh' indicates that the word in which it is placed had a guttural pronunciation.}\\
$> ... <$	&\multicolumn{2}{p{130mm}|}{`More than' and `less than' signs indicate that the talk they encompass was produced noticeably quicker than the surrounding talk.}\\
$\leftarrow$	&\multicolumn{2}{p{130mm}|}{Arrows in the right margin point to specific parts of an extract discussed in the text.}\\
\protect[LDC301]	&\multicolumn{2}{p{130mm}|}{Extract headings refer to the transcript library source of the collected data.}\\
\hline
\end{tabular}
\caption{ Gail Jefferson's transcription scheme as described in Hutchby and Woolffitt~\cite{HutWoo98}} \label{GJ}
\end{table}

Many practitioners have provided a general description of how to do CA,
including ten Have~\cite{tHave99}, Hutchby and Wooffitt~\cite{HutWoo98},
Pomerantz and Fehr~\cite{PomFehr97} and Seedhouse~\cite{Seedhouse04}.
Of these, Paul Seedhouse has a strong focus on action sequences and offers
the following account of the analysis process:
\begin{enumerate}
\item  Locate an action sequence or sequences.
\item  Characterise the actions in the sequence.  This might look like
form-function matching, speech act analysis, or discourse analysis depending
on one's background.
\item  Examine the action sequence(s) in terms of the organisation of turn taking, focusing especially on any disturbances in the working of the system.
\item  Examine the action sequence(s) in terms of organisation.  Not only
adjacency pairs and preference organisation, but more widely at any action undertaken in response to other actions.
\item Look at repair in action sequences.
\item How do the speakers package their actions and what is the significance of this.
\item Uncover any roles, identities or relationships which emerge in the details of the interaction.
\end{enumerate}

This is basically the approach taken here.

\subsection{Norms, preference, and giving account}
In CA there has been much made of the realisation that questions are usually
followed by answers, greetings by another greeting and so on.  The first
part `brings on' the second part of an adjacency pair. 
This behaviour is `normative' rather than a rule and as such the second
part is not guaranteed to occur.  As an example of normative action,
consider what happens when someone buys a round in the pub.  They
expect that, at some point in the future, the others will also buy him or
her a drink.   The social act of buying a round sets up expectations
both in the actor, the actees, and observers. 
For someone to buy a reciprocal round is normal behaviour and hence goes
unnoticed.  To \emph{not} reciprocate is not only noticed, but
interpreted.  If no explanation is forthcoming, then the action may
indeed bring on sanctions of some kind.   This is the terminology of
the latter work in CA - adjacency pairs, noticing and sanctions - and this
normative behaviour forms the basis of language as social action.  By
participating in a conversation, one sets up a context in which certain
actions are normal.   To act in accordance with these norms is to
play by the rules and such action is `seen but unnoticed'.   
Questions are answered and greetings are greeted.
As a social actor one can indeed deviate and answer a question with another
question, but such an action is noticed and as
such is interpreted by one's conversational partner as meaningful. 
Consider an example from Eggins and Slade~\cite[page 29]{EggSla97}:
\begin{quote}
Consider, for example, the two turns at talk below:

\begin{tabular}{rl}
\textbf{A} & What's that floating in the wine? \\
\textbf{B} & There aren't any other solutions. \\
\end{tabular}

You will try very hard to find a way of interpreting B's turn as somehow
an answer to A's question, even though there is no obvious link between them,
apart from their appearance in sequence.  Perhaps you will have decided
that B took a common solution to a resistant wine cork and poked it through
into the bottle, and it was floating in the wine.  Whatever explanation
you came up with, it is unlikely that you looked at the example and simply
said ``it doesn't make sense'', so strong is the implication that adjacent
turns relate to each other.
\end{quote}
The notion of sequential relevance is so strong in humans that entire
conversations can follow and still be seen as relevant.  Consider
this example from Mann~\cite{Mann88} when he is introducing the idea of
`dialogue games':
\begin{quote}
\begin{tabular}{rl}
\textbf{Child} & I'm hungry. \\
\textbf{Mother} & Did you do a good job of your geography homework? \\
\end{tabular}
\end{quote}
In this case Mother is effectively saying you can have something to eat
after you have done your homework.  She has ignored the child's 
statement; but that in itself is a speech act \emph{because} it is
interpreted as being sequentially relevant. 
Of course conversation does require that some things are not sequentially
relevant, but as Eggins and Slade go on to point out:
\begin{quote}
... if a speaker does \emph{not} wish an utterance to be interpreted as
related to immediately preceding talk, s/he needs to state that explicitly,
using expressions such as \emph{To change the subject ...}, or \emph{By the
way...}
\end{quote}
To answer a question is to do what is expected; to \emph{not} answer the
question is interpreted as relevant to the question unless it is explicitly
flagged.   As seen below, humans do this kind of behaviour even when
talking to a machine.

Questions are (normally) followed by answers; greetings (normally)
followed by greetings, and anything else is interpreted as somehow
relevant unless explicitly marked otherwise.  But within this normative
behaviour, there are still preferred responses.  This preference
is bought on by social convention and by the construction of the
question.  Usually for a yes/no question this is the
affirmative.  To ask ``You want to go to the pub?'' is different to
asking ``You don't want to go to the pub today do you?''. The preferred
response brings on short answers, while a non preferred response brings
on justifications, apologies and explanations.  Again from
Seedhouse's book,
\begin{quote}
Dis-preferred responses are generally accompanied by hesitation and
delay and are often prefaced by markers such as \emph{well} and
\emph{uh} as well as by positive comments and appreciations such as
``You're very kind.''  They are frequently mitigated in some way
and accounted for by an explanation or excuse of some kind. 
A's turn in Extract 1.8 exemplifies all of these phenomena:

\begin{tabular}{rp{120mm}}
\textbf{B:} & uh if you'd care to come over and visit a little while this
morning I'll give you a cup of coffee.\\
\textbf{A:} & hehh well that 's awfully sweet of you, I don't think I
can make it this morning. hh uhm I'm running an ad in the paper and  -
and uh I have to stay near the phone.  [Atkinson \& Drew, 1979, p. 58]\\
\end{tabular}
\end{quote}

The proposal is that conversational machines are fine when things go
as expected - we know how to make conversational agents that can
act normatively - but when things go wrong, conversational machines
fail to account for their misbehaviour.  This leads to verbal abuse.
From an instance of verbal abuse (the sanction) we can
work back through the text to see how the system has failed to account for a
dis-preferred response.

\section{The Communicator Project} \label{communicator}

The goal of the DARPA Communicator program was 
\begin{quote}
to develop and demonstrate ``dialogue interaction''
technology that enables [people] to talk with computers, such that
information will be accessible [...] without ever having to touch a
keyboard.   The Communicator Platform will be wireless and mobile,
and will function in a networked environment.  Software enabling
dialogue interaction will automatically focus on the context of a dialogue
to improve performance, and the system will be capable of automatically
adapting to new topics so conversation is natural and efficient.
\cite{CommLDC}.
\end{quote}
In a military context the aim was ``command and control to the edge'' -- the
idea that the people on the ground had direct real time access to information
and directions, while still being able to do the job they were there to do.
 In general keyboards are fine for office workers, but for almost everyone
else, they are a distraction.  Truck drivers are not going to use a keyboard
to access sat-nav equipment; firefighters are too focused to use a PDA, and
nurses bare the brunt of inappropriate technology in hospitals.   The ICT
revolution is not reaching the people who actually do things.

The Communicator evaluation tasks were based around the notion of a
user driving along the highway and needing to use their hands free mobile
phone to arrange flights, hotels and hire cars.  Could someone reliably
use a spoken language interface, in a seamless and easy way, to arrange
these things while doing something else?
The answer seems to have been less than positive.
As part of the project, participants in the evaluation were given a
survey and one of the questions was about future use.  Figure 1 shows
that quite a few callers were less than impressed.

\vspace{3mm}
\begin{tabular}{|r|c|}
\hline
\multicolumn{2}{|p{80mm}|}{
Based on my experience in this conversation
using this system to get travel information, I would like to use this
system regularly.} \\ \hline
agree completely: & 170 \\
mostly agree: & 325 \\
no opinion: & 196 \\
mostly disagree: & 233 \\
disagree completely: & 390 \\
(did no complete) & (37) \\
\hline
\end{tabular}
\vspace{3mm}

\noindent
So what is happening?  Why do so many people \emph{not} want to
use the system regularly? 
The LDC Communicator corpus contains transcripts for 1684 phone calls
made in 2001 to various systems from, in alphabetical order, AT\&T,
BBN, Colorado, CMU, IBM, Lucent Bell Labs, MIT and SRI.  
For nearly all these, the sound files are also available
of the caller's side of the conversation.

\subsection{Using CA on the Communicator transcripts}
As a piece of qualitative research, the aim is to
identify a \emph{species} of social phenomena that lead
to people not wanting to use the system again.  As discussed above,
the sampling should indeed have `maximal coverage' - one is discovering
a class of phenomena about which one can say something interesting
and ideally, that class will have well defined edges.

The first step was to look for swear words based on the assumption that
these are the sanction bought on by a failure of the system to account
for its actions.  Unfortunately there were only three calls out of the
1500 or so calls that used uniquely identifiable swear words.  Given
the figure of 2\% for the AMITI\'{E}S project mentioned above, we would
expect around thirty calls in the Communicator data to involve 
identifiable abuse.  Where are they?  Why were the participants all
so well behaved?  Given the survey results, it was obviously not because
the system was perfect.  It is perhaps because Americans are more
polite, but it seems more likely that the participants were self
conscious of the fact they were being watched.  It would perhaps be
useful if in the future the guidelines for the participants were made
available along with the data. 

The next step was to look for other indications of sanction.
Looking at the supplied transcripts I noticed
that several callers use the expression `start over'.  This is not
something that occurs regularly in `natural' conversation and suggests
the caller has failed to make natural language work for them. 
There are 239 calls that use this phrase at least once, with one call
containing 13 ``start over''s and three containing 6. 
Working through the LDC transcripts for the 239 calls of interest, failed
ASR is certainly a big factor in callers deciding to start over - particularly
if they had only just started and felt it was easier to start again
than go through a step by step process of correcting the data.  
As can be seen below, there is a need for better ASR, but the more interesting
cases for me are those cases where something else has caused the `trouble'.

The first call of interest was chosen because the caller expresses
frustration and, if I were her, I would have been tempted to swear.
In addition, although there
is a problem with the ASR, the call makes the same mistake as the system
and as such the unhappy outcome cannot be explained away as a need for
better ASR.  Transcripts of the call was made using the
publicly available Audacity software~\cite{audacity}, and
Jefferson's transcription scheme shown in Figure~\ref{GJ}.
Note that neither this call, nor the one below, involved overlap. 
Although many calls to the Communicator
systems did involve overlap and the gross timings of overlap are
available in the log files, the recording scheme recorded the caller and the
system as separate data files and so listening to the overlap as it happened
would be quite difficult.   In any case, the callers know that they are
talking to a machine and so tend to be more strict about turn taking than
two humans engaged in natural talk.
Finally, note that in this call some lines appear to be transposed in the
transcript provided on the DVD.  This has been corrected in the
transcript presented here: 

  60941\_XX\_XX\_04\_20010608\_062510.xml

\begin{tabular}{rrl}
&    Sys:&please remember to speak after the tone. if you get confused at \\
&	  &any point you can say start over to cancel your current itinerary. \\
&	  &are you calling from a cellular phone? \\
&  000 Usr:&n$\downarrow$o \\
&      Sys:&what are your travel plans? \\
&  001 Usr:&i'd like t' go from \underline{duluth::}(0.3)minnesota(0.4)to \\
&	  &chicago illinoi\underline{s} \\
&      Sys:&what month and day do you wish to depart toulouse \\
$1\rightarrow$& 002 Usr:&november seco$\downarrow$nd(0.4) twenty s- \\
&      Sys:&what time do you want to leave toulouse       \\
$2\rightarrow$& 003 Usr:&((background)) \\
&      Sys:&what time do you want to leave toulouse \\
&  004 Usr:&in the late morning \\
&      Sys:&if you have a preferred airline please state it now or say \\
&	  &no preference \\
&  005 Usr:&american airlines \\
&      Sys:&let's see then    i have you going from toulouse minnesota \\
&	  &to chicago illinois on november second is that correct? \\
$3\rightarrow$&  006 Usr:&no \\
&      Sys:&oops      let's try again. i have you departing from \\
&          &toulouse minnesota. did i get that right? \\
&  007 Usr:&yes! \\
&      Sys:&i have you arriving in chicago illinois \\
&          & have i got that right? \\
&  008 Usr:&yes \\
&      Sys:&you will depart on november second is that correct? \\
&  009 Usr:&No \\
&      Sys:&what month and day do you wish to depart toulouse \\
&  010 Usr:&november(0.2)twenty nin$\downarrow$th \\
&      Sys:&let's see then.   i have you going from toulouse minnesota \\
&	  &to chicago illinois on november twenty ninth is that correct? \\
&  011 Usr:&yes \\
&      Sys:&please let me check availability and see what i can find \\
&      Sys:&no records satisfy your request. would you like to depart \\
&	  &on a different date?                           \\
\end{tabular}

\begin{tabular}{rrl}
$4\rightarrow$&  013 Usr:&n$\uparrow$o! \\
&      Sys:&would you like to start over? \\
&  014 Usr:&yes \\
&      Sys:&what are your travel plans \\
&  015 Usr:&duluth minnesota(0.2) to chicago illinois \\
&      Sys:&what month and day do you wish to depart toulouse \\
&  016 Usr:&november \underline{twenty} n$\downarrow$inth \\
&	  &-- call contiues -- \\
\end{tabular}

Having identified something of interest and transcribed it, the next
phase is analysis.  This starts by identifying actions, and then
moving on to explain why particular actions appear where they do. 
The action of particular interest is user utterance 13 when the caller
says ``n$\uparrow$o!''  At turn 13 the user's intonation goes up which
- using my common sense understanding of English usage - suggests the
caller is indignant. 
Listening to the tape one can imagine that, under non-test conditions,
the caller would have followed ``n$\uparrow$o!'' with ``I do not want to
depart on a different date!'' and perhaps ``Idiot!'' or something similar.

Before looking at the actions used here, note how one's choice of actions
is not theory neutral.  In this transcript the caller says `no' four times
as a stand alone utterance and in each case there are noticeable differences.
Whether or not one interprets these variations as significant depends on
one's interests.  For instance, the intonation goes down at the end of the
first `no' and this could be interpreted as a certain amount of nervousness
at the start of the call, or it could be interpreted as an emphatic statement
about something she is absolutely certain of.   If one was interested in
attitudes to automated call handling systems, then presumably one might
consider the intonational shift as a form of action.
In this paper however, the interest is in the sequence of actions that
led to the caller being indignant.

Note that the mark-up is not about the semantics of the text, but about who
does what \emph{as interpreted by the other person}.  I can
tell this by using my commonsense understanding as an expert language user,
not as an expert on theoretical linguistics.
The system gives instructions, asks questions and confirms it's beliefs.
The caller answers questions and, involuntarily, hesitates and expresses annoyance.
The system very clearly has control of the dialogue structure.
The system has the initiative at all times, with the caller simply answering the
questions.  The interesting cases -- the events that deviate from the strict
question-answer format -- are numbered and start with the caller at 1 trying to
do a self-repair.  She has said November second, when she meant to say twenty
second as evidenced further down the transcript.  At 2 the system has asked a
question and the caller doesn't say anything.  The system then repeats the
question and the caller provides an answer.  At 3 we find out (as does the
system) that the system has the wrong information.  This prompts a sub-dialogue
that asks the caller to confirm or correct the data the system has.  Finally,
just before 4, the system looks for a flight that fits the callers requirements
and finds there are none.
At this point, the system offers some help and asks if the caller would like to
fly on a different day and at 4 the caller, \emph{in addition to answering the
question}, is indignant.  How did that come about?  

Having decided which actions are relevant, the next step is to use commonsense
to explain why particular actions took place.  In other words to address the
question of why this, in this way, right here?  Embracing the ethnomethodological
principle that my common sense is an appropriate tool for studying actions
at the social level, why does the caller sound annoyed? 

The caller has rung a travel booking service with the expectation that it
can book her travel.  When it can't, she is surprised.  She has,
no doubt, flown from Duluth to Chicago before, and \emph{doesn't believe}
there are no flights on the twenty ninth.  This, in itself, is not so
bad - this type of miss-understanding is common in human conversation -
as evidenced by the fact that the human has made the same ASR error
as the machine.  The \emph{problem}, according to my common sense
as an expert language user, is what the system does next.  
At (4) the system provides a piece of information - that there are no
records that satisfy the caller's request - and in the same turn it goes
on to make a suggestion - perhaps a different day.
Although this response makes perfect sense from the perspective of information
state update~\cite{KreMat00}, from a social actor it is rude.  If it does not appear
rude to the reader, consider how you would feel if a call centre operator used
these words.  From a machine we expect information but from a human, we expect
assistance.

In a classic text on politeness, Brown and Levinson~\cite{BnL} describe
how socially responsible actors avoid placing restrictions on their
conversational partner and will go to some lengths to avoid such actions.
In this case their explanation would be that the system has committed an
FTA - a face threatening act - and made no attempt to mitigate it.
In this case the system has `restricted the caller's options' by asking
if she wants to change the date.  When she turns this down, it further
restricts her options by asking if she wants to start-over.  She is
indignant because the system was rude.
The notion of an FTA is a theoretical construct that has, for some
reason, become an accepted concept in the CA literature. 
As discussed elsewhere \cite{wmod01}, FTAs are not part of commonsense
reasoning used by members of the community and as such, it is hard to see
how an explanation in terms of FTAs can be
seen as ethnomethodological in nature.  Indeed explaining the events
in terms of politeness does not help because politeness can only be
described in terms of events such as these.

An alternative description - more in keeping perhaps with CA-EM in general
is that ``no records satisfy your request'' is a dis-preferred response and
as such must be mitigated in some way.  In human-human conversation we might
expect ``hesitation and delay, ... prefaced by markers ... as well as positive
comments and appreciations. ...  mitigated in some way and accounted for
by an explanation or excuse of some kind''~\cite[page 24]{Seedhouse04}.
Her indignant reaction is \emph{sanctioning} the system for not following the
norms of civil society and \emph{accounting for} the dis-preferred response saying
that the system could not help her with a flight.

Before moving on to look at another transcript, note that there are
many more things one might want to notice in this transcript. 
Following the Seedhouse description of doing CA in section 4, I should
look for repair, the packaging of actions, and note roles, identities or
relationships that emerge in the text.  Indeed the caller attempts
a repair when she makes a mistake with the date at (1) but the system
does not let her do it.  The hesitation at (2) is another indication
that something is wrong which the system ignores. 
Indeed the systems response to this hesitation is again quite rude.
In a way, the system is making a claim
about its social status with this act - it is something that one might
accept from a quiz master but it is not something one would accept
from the young man on the cash register at the supermarket.
As De Angeli et al point out~\cite{ang01}, the role adopted by a system is
often a role the system designer might adopt whereas the perception
of a system user can be quite different.  In this paper the focus
is on the actions leading to events at (4) but there are plenty of
issues raised for automated call handling in this one transcript.
The obvious question is then, how
should the system respond given there are no records?  In what
follows we look at a call that can offer some guidance.

\subsection{Transcript 2: sub-dialogues}
In the previous transcript the system was in control, and the user
stuck strictly to answered the questions.  In this case the caller
takes the initiative.    In both transcripts the system demonstrates
the acceptance of un-called for but relevant information; in this transcript
the user also tries to introduce his own topic.

 --- 10359\_2\_02\_02\_20010524\_082717.xml ---

\begin{tabular}{rrl}
&      Sys:& what trip would you like to take? \\
&  001 Usr:& leaving at$<$lan$>$ta flying t' lon$\uparrow$don \\
&      Sys:& flying from atlanta to london.. \\
&      Sys:& what day are you leaving atlanta? \\
&  002 Usr:& september twenty fou$\uparrow$rth \\
&      Sys:& on monday, september twenty fourth.. \\
&      Sys:& about what time do you want to leave? \\
&  003 Usr:& early even$\uparrow$ing \\
&      Sys:& leaving in the early evening.. \\
&      Sys:& please stand by while i look up the flight schedules on the \\
&	   & web. \\
&      Sys:& ok, got them. \\
&      Sys:& i can only get you a flight in the early evening on united. \\
&	   &would that be ok? \\
$1\rightarrow$& 04 Usr:& i pre\underline{fer} british airways \\
&      Sys:& there are no flights on british airways. \\
&      Sys:& i have just one flight. \\
&      Sys:& the flight is on united at six fifty five pm, arriving on \\
&	   &tuesday, september twenty fifth at nine forty eh m. \\
&      Sys:& is that ok? \\
&  005 Usr:& n$\downarrow$o \\
&      Sys:& if you'd like an earlier or later flight, or a different \\
&	   &airline, please let me know \\
&      Sys:& the flight is on united at six fifty five pm, arriving on \\
&	   &tuesday, september twenty fifth at nine forty eh m. \\
&      Sys:& is that ok? \\
&  006 Usr:& how about an earlier flight(.)on british air$\downarrow$ways \\
&      Sys:& leaving before six fifty five pm instead. there are no \\
&	   &flights on british airways. \\
&      Sys:& i have 14 flights. \\
&      Sys:& the first flight is on continental at four forty five pm, \\
&	   &arriving on tuesday, september twenty fifth at eight thirty \\
&	   &eh m, with a connection in newark. is that ok? \\
&  007 Usr:& \underline{n$\downarrow$o} \\
$2\rightarrow$ & & (0.2) do you have \underline{any}thing on british airways \\
&      Sys:& there are no flights on british airways.         \\
&      Sys:& the second flight is on delta at four pm, arriving on \\
&	   &tuesday, september twenty fifth at seven forty five \\
&	   &eh m, with a connection in newark. is that ok? \\
$3\rightarrow$& 008 Usr:& \underline{n$\downarrow$o} give me the ori\underline{g}inal \\
&      Sys:& the third flight is on air canada at three fifty seven pm, \\
&	   &arriving on tuesday, september twenty fifth at six fifty \\
&	   &eh m, with a connection in toronto. is that ok? \\
&  009 Usr:& \underline{N$\downarrow$o} \\
\end{tabular}

\begin{tabular}{rrl}
&      Sys:& the fourth flight is on united at three fifty seven pm, \\
&	   &arriving on tuesday, september twenty fifth at six fifty \\
&	   &eh m, with a connection in toronto. is that ok? \\
$4\rightarrow$& 010 Usr:& $>$go back$<$ \\
&      Sys:& sorry, i'll go back to the question i was asking before. \\
&      Sys:& the third flight is on air canada at three fifty seven pm, \\
&	   &arriving on tuesday, september twenty fifth at six fifty \\
&	   &eh m, with a connection in toronto. is that ok? \\
&  011 Usr:& $>$start over$<$; \\
&      Sys:& sorry, i didn't understand that. \\
&      Sys:& the third flight is on air canada at three fifty seven pm, \\
&	   &arriving on tuesday, september twenty fifth at six fifty \\
&	   &eh m, with a connection in toronto. is that ok? \\
$5\rightarrow$&  012 Usr:& \underline{start}(0.2)ov$\downarrow$er \\
&	    \\
&	  & -- call continues -- \\
&	    \\
\end{tabular}

On first reading the system is very pushy.   It offers the first flight
three times, and then when it gets a list of 14 possible flights, it insists
on working through them.  Note that once again there is an ASR problem
when the caller runs ``start'' and ``over'' together at utterance 11,
but that is corrected at 12 where he separates them with a fifth of a 
second gap.  This gap is bigger than necessary and, listening to the
recording, one can almost hear the silent anglo-saxon in-fix modifier\footnote{In English modifiers go at the beginning or end of a word with the exception of swear words which can be inserted in the middle - ``abso-bloody-lutely'' being a mild example.}.
The caller is not ``like to use this system regularly'' and the question
is, how did the two of them reach that situation?

In the previous transcript the caller stuck to answering the system's
questions but in this case, the caller takes the initiative. 
At (1) the system asks a yes/no question, and the caller's response is
to add a preference for British Airways.  This is of course not the
normative response to a yes/no question - his behaviour is not the expected
second part of the adjacency pair and as such, it is noticeable,
accountable, and possibly sanctionable. 
So why is the user saying this, in this way, right here? 
As an action at the social level, the user is expressing a \emph{preference}
for British Airways.   The evidence for this is semantic
(``I prefer British Airways'') but also expressed in the way this modification
to the requirements is positioned right here.  The caller has stated
this preference in response to a yes/no question and as such the caller's
action must be, by the rules of sequential relevance, interpreted as relevant
to the question.   Like the mother and child dialogue discussed above,
the answer to the system's yes/no question depends on how this new topic
pans out.  Preempting the following discussion, the answer to this
question is ``yes,'' and it occurs at (3).  We have here an example of a
`higher level structure' that would seem at odds with the term
`adjacency pair' but is in keeping with the current day interpretation of
conversation analysis.

When the caller expresses a preference for BA flights in response to the
system's yes/no question however, the system picks up on the semantics. 
The system proceeds to update its information state, run the new query,
finds there are no BA flights and reports this fact. 
As there are no BA flights, the system again offers the United Flight, this
time with more detail. 

This move appears `pushy' to me as a member of this language using community
and, presumably, to the reader as a member of that same community.
Why is that?  `Obviously' it is because the system is sticking to its plan
to give the user the United flight it has already found. 
The system's actions are not those expected, but that is fine because the
action can be \emph{accounted for} in that the system is of
a pushy nature.  Whereas the user expected to
be given the option of a BA flight, the system didn't offer one, and didn't
try hard enough to do so because the system `wants' to sell the flight it has.
This pushiness is fine as far at the user is concerned - he has dealt with
pushy sales people before.  The caller's response is to simply say ``no'' to the
offer at utterance 5.  From a social perspective it is now the system's
responsibility to move the conversation forward.  What should the system do?
 Where can the conversation go when the system has no more flights to offer?
Note this is exactly the same situation as the ``no flights satisfy your
request'' situation in transcript one. 
In this case however the system's response is far more polite -- in fact it is not
so much polite as simply \emph{seen but unnoticed}. 
 The system's response is to suggest that the user try an earlier flight or
 a different airline.  In this
case the system leaves open the caller's options.  In the previous
case the system asked if the caller would like to change the date; in this
case the system says that, if he would like an earlier flight, then he is
at liberty to change the time of the flight; or perhaps a different
airline?   There is a suggestion that the user has many options, and
these are but a few.  The system's strategy in this case is far more in
keeping with the norms of civil conversation. 

The user's response (2) is to ask ``Do you have \emph{any}thing on british
airways?''  Looking at this move as social action and asking the question
why this, in this way, right here, the user is not looking for a flight to
book at this point.  Would he accept \emph{any} BA flight?  Not
likely - the question of the original United flight is still open - he is
still considering it.  The user thinks he is talking about his preferred
option of flying BA; the system however (behaves as if it) is thinking it is
looking for a flight.  The system picks up on the information content,
and notes that the user has changed both the time \emph{and} the airline;
the system treats it as an `or' expression and finds nothing on BA and
another 14 flights. 

So why is the caller asking for \emph{any}thing on BA if he does not want
just any BA flight?   As in the previous transcript, it seems he
\emph{doesn't believe} there are no BA flights.  What he seems to
be doing is to explore the database.
In this case at least the user, when faced with ``no records match your
request'', wants to find out why.  
Between utterance (1) ``I would prefer british airways'' and (3) ``give me the
original'' the caller 'puts on hold' his interest in booking a flight and
attempts something else.  He attempts to understand why this service doesn't
have any BA flights to London.  The \emph{strategy}
he uses is to try a series of explicit queries that make sense only in
the context of his understanding of how the world works.  In effect he is
using his own commonsense to help out. 

At (3) the user gives up and asks for `the original'. 
He has stopped exploring the database and returns to the system's topic,
namely booking flights.  
 Using common sense he wants the United flight leaving at six fifty five,
but how does one's common sense tell us this?  One could figure
it out from the semantics - the verb `give' has a preference for
givable things and the things being given in this context are flights. 
There are however two possibilities here - the United six fifty five, or the
first from the list of 14 - the Continental flight at four forty five. 
To me - and presumably to the reader and the user - there is no question the
requested flight is the United.  Looking at action as social rather
than semantic, the reason the preference is for the United flight is that the
question prior to (1) is still open and this must be, in some way, an answer
to that question. 
The system misinterprets the user's dialogue move here - in fact doesn't
understand this at all; it picks up on the ``no'' - and offers the next flight
on its list.  At (4) the user asks the system to ``go back''. 
In the user's head the place to go back to is obvious but the system
interprets this as the previous question.  That is not what the caller
meant and he gives up, but now he is getting angry and the system doesn't
recognise his ``start over'' request.   The caller in this case must be
admired for his calm.

\section{Analysis and a Proposal} \label{RMI}

What should an automated service provider do when, using the example from the
first transcript, ``no records satisfy the request''?  
In the first transcript the system makes a series of suggestions that
ends with the suggestion to start again --- a particularly frustrating
response that, in less monitored circumstances, is likely to prompt the
user to verbal abuse.  From the CA perspective the problem is
\emph{not} that the system doesn't have any  flights from
Toulouse to Chicago, but that saying there is nothing is a
dispreferred response and as such we humans expect the response to be
 accompanied with ``hesitation and delay, ... prefaced with markers
such as \emph{well} and \emph{uh} ...''.

Interestingly, the second call has exactly the same problem but, in
that case, the invitation to change some of the parameters is made in
such a way that it appears to be opening up options rather than
restricting them.  In the second case the response is not a face
threatening act, and thus goes seen but unnoticed.  Unfortunately the
user just doesn't believe that there are no British Airways flights to
London - presumably the system only listed American owned airlines -
and the system continues to sell the flights it has. 
This 'pushiness' accounts for the system's dispreferred response but,
unfortunately from the evaluation perspective, the user has dealt with
pushy people before and tries the same technique on the computer.
What the user ends up doing is to explore the data-base.

In this second transcript the user introduces his own topic and
expects the system to follow.  The computer can answer questions such
as ``do you have any BA flights to London'' - the problem is that the
system sees this question as related to the current task of booking flights.
What is needed is some notion of higher level structure, and being
able to recognise when to drop a topic, or at least put it on hold.
If the system had such a strategy for helping the caller explore the database,
then the strategy might be offered to the user.  In the first call, rather
than saying ``Would you like to fly on a different date?'' the system might have
said ``It appears the only flights to Toulouse go via London or Paris'' which
would have alerted the caller to the real issue, namely the Toulouse/Duluth ASR
problem.

But having such a strategy is not necessary.  If the system had said ``There are
no records which match your request.  Umm'' The caller is not going to start
talking about her weekend; if she introduces a topic, it will be relevant to
the process of booking flights.  Giving the caller the initiative would allow
her to user her knowledge of the world to help the system find a solution for
her. The only issue is then to notice when the user returns to the
topic of booking a flight, and in this second transcript, that is
clearly marked with the ``give me the original.''

A system that knows when and how to introduce topics, and that can follow when
the user introduces new topics, is what is commonly known as a
\emph{mixed initiative} system.  The term however has been appropriated by
the commercial interactive voice response companies to mean a system that can pick up required
information any time the user offers it.  For example picking up the date when
the system asks for a destination, and the user responds with a location and
a date.  Others interested in topic shift have called this `mixed initiative
at the discourse level'~\cite{MaxGuide}, however the term `mixed
initiative' is less of a mouth full. 

It has often been assumed that a dialogue system that could handle
mixed initiative would need a huge range of world knowledge in order
to take up arbitrary topics introduced by the user.
However a human talking to a travel agent is not going to randomly
introduce the topic of quantum physics, and if they do, they will need
to account for the shift. Without an acceptable accounting for, the
agent can justifiably impose a sanction -- probably by threatening to
terminate the conversation.

By providing for mixed initiative dialogue, the human can use his or
her common-sense
to find out why there was dispreferred response and thus account for
the failings of the system.  Hopefully understanding why will lead to
repair, be they semantic (BA flights aren't listed) or speech recognition
(the Toulouse/Duluth distinction) between the user and their helpful
artificial conversational partner.

\section{Conclusion}

Not as a scientist, but as a practising member of the language using
community, I have \emph{direct} access to the what-is-being-done of
language in use.  Although there was surprisingly little swearing in
the Communicator transcripts, the survey showed there was surprisingly
high levels of user dissatisfaction, and listening to the recorded
speech I can tell when people become frustrated, annoyed, and indeed
angry.  Recordings of the above conversations are available through
the LDC and, as part of the CA-EM methodology, the research community
is invited to check the claims made here using their own commonsense
understanding of language in use.  Whereas corpus analysis provides
data about average or normal behaviour, the exceptions are a different
matter.  CA provides the techniques to see through the commonsense
of everyday language in use, to make explicit the actions of speakers,
and hence to make possible computer models of what people actually do
with language.

The problem is to figure out why they have become annoyed.
The notion that a response is either `seen but unnoticed', or can
be `accounted for,' or the responder `risks sanction,' suggests
that working backward from a sanction, we will find a failure to
account.  Using a conversational agent to access a data-base, the
preferred response is for the agent to provide a solution; when a
system can find "no records that match your request," it is not enough
to simply say that that is the case and terminate the conversation.
The agent needs to make suggestions \emph{without being seen to limit
options}, to accompany the dis-preferred response with ``hesitation
and delay, ... prefaced with markers such as \emph{well} and \emph{uh}
...'' and/or hand the initiative to the requester.

Of course there is no need for a conversational agent to handle mixed
initiative dialogues when things go well and many working on language
technologies assume that it is the mistakes that annoy users.  However
humans make mistakes all the time~\cite{wmod01} but having the ability
to introduce topics allows us humans to repair the conversation.
Rather than putting more effort into reducing the problems caused by
speech recognition errors for example, a better solution is perhaps to
let the errors happen, but enable the system to repair them.
The number of problems to which really mixed initiative systems is
a solution is an open research issue.

\section{Thanks and Support}
A version of this paper was published in \emph{Interaction Studies}, vol 9:3
pp.434--457, December 2008, doi https://doi.org/10.1075/is.9.3.05wal. The work
was in part funded by a grant from the Engineering and Physical Sciences
Research Council (EPSRC) in the UK, EP/F067631/1, 2008 ``Engineering Natural
Language Interfaces: can CA help?''.

\bibliographystyle{plain}

\end{document}